# Word level Script Identification from *Bangla* and *Devanagri* Handwritten Texts mixed with Roman Script

Ram Sarkar, Nibaran Das, Subhadip Basu, Mahantapas Kundu,
Mita Nasipuri and Dipak Kumar Basu

**Abstract**— India is a multi-lingual country where Roman script is often used alongside different Indic scripts in a text document. To develop a script specific handwritten Optical Character Recognition (OCR) system, it is therefore necessary to identify the scripts of handwritten text correctly. In this paper, we present a system, which automatically separates the scripts of handwritten words from a document, written in *Bangla* or *Devanagri* mixed with Roman scripts. In this script separation technique, we first, extract the text lines and words from document pages using a script independent Neighboring Component Analysis technique [1]. Then we have designed a Multi Layer Perceptron (MLP) based classifier for script separation, trained with 8 different word-level holistic features. Two equal sized datasets, one with *Bangla* and Roman scripts and the other with *Devanagri* and Roman scripts, are prepared for the system evaluation. On respective independent text samples, word-level script identification accuracies of 99.29% and 98.43% are achieved.

———————————— ◆ ————————————

## 1 INTRODUCTION

TO develop a successful multi-lingual OCR system, separation or identification of different scripts is a very important step. In a multi-lingual country like India, it is an utmost essential for designing an OCR system. India has more than 22 official languages and 12 different scripts are used for these languages. Moreover, English is taught and used largely almost allover India. The colonial past has a great influence on the Indian culture. As a consequence, English text is widely used in many Indian script handwritten/printed documents leading to difficulty in determination of the script(s) of the document image. Most of the published methodologies [4,6,12-18] on automatic scripts separation, discussed about printed text document. A few number of research work [5,7-8,10-11,18] are available which applied on handwritten text document.

The above-mentioned reasons, combined with the increasing relevance for multi-script handwriting recognition, motivated us to develop an automatic word level script separation technique for handwritten English (Roman) script mixed with either *Bangla* or *Devanagri* scripts.

- *R.Sarkar is with the Computer Science and Engineering Department, Jadavpur University, Kolkata, India.*
- *N. Das is with the Computer Science and Engineering Department, Jadavpur University, Kolkata, India.*
- *S. Basu is with the Computer Science and Engineering Department, Jadavpur University, Kolkata, India.*
- *M.Kundu is with the Computer Science and Engineering Department, Jadavpur University, Kolkata, India.*
- *M. Nasipuri is with the Computer Science and Engineering Department, Jadavpur University, Kolkata, India.*
- *D. K. Basu is AICTE Emeritus Fellow, Computer Science and Engineering Department, Jadavpur University, Kolkata, India.*

Among the relevant research contributions, the technique in [5] was based on analysis of connected component profiles extracted from the destination address block images. It did not emphasis on the information provided by individual characters themselves and does not require any character/line segmentation. In [7], a system was developed by Hochberg et al. to automatically identify the six different scripts (Arabic, Chinese, Cyrillic, *Devanagri*, Japanese and Roman) used in handwritten document. First, they identified connected components assuming eight-connectedness. Then five features were extracted from all the connected components. Mean, standard deviation and skew for each component feature across all components were calculated. A linear discriminant analysis was trained to classify the script of the new documents, and tested using writer-sensitive cross-validation. They used a similar procedure to identify the language. In [8], Roy et al. developed a technique for script separation of handwritten postal document of *Bangla*, Roman and *Devanagri* scripts. They used Run Length Smoothing Algorithm (RLSA) to segment the document pages into lines and then into words. Then using fractal-based, busy-zone and topological features and a Neural Network (NN) classifier they identified the scripts. In [10] authors had proposed a script separation technique of Roman and Oriya scripts for Indian Postal automation. Authors used a piecewise projection method to extract the lines and words. Finally they used a NN classifier, trained with different features like, water reservoir based features, fractal dimension based features, topological features etc., to identify the scripts. In [11], Dhandra et al. used two-stage approach for script identification in handwritten documents. First, they applied some global and local features to identify the text words. Then in the second phase, they identified the numeral written in different scripts. To test the system, they used Kanada, *Devanagri* and Roman scripted handwritten document. A method for Arabic and Latin text block differentiation for both printed and handwritten scripts is proposed in [18]. This method was based on a morphological analysis for each script at the text block level and a geometrical analysis at the line and the connected component level.

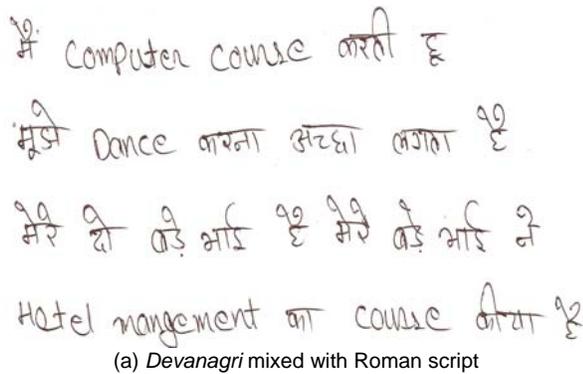

(a) *Devanagri* mixed with Roman script

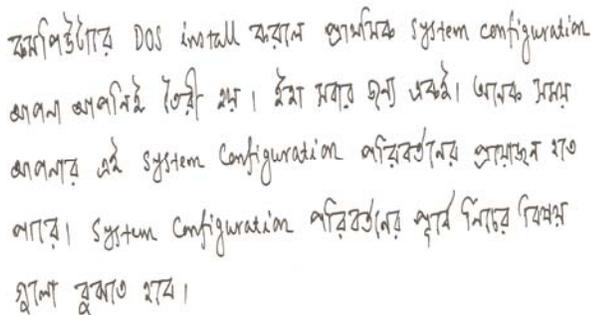

(b) *Bangla* mixed with Roman script

Fig. 1 (a-b): Sample of Multi-script documents

Despite these research contributions few work [5,8,10,11] is available in the literature on word-level script identification from handwritten Indic script documents mixed with Roman script. It may be worth mentioning in this context that script identification techniques for a complete document page or even a complete text line does not work for multi-script document pages. This is due to the presence of Roman script words in between many text lines in a multi-script handwritten document page. This is illustrated in Fig. 1(a-b). Therefore we need a word-level script identification algorithm for effective character segmentation and subsequent recognition. The techniques designed for this purpose is described in the following section.

## 2 PRESENT WORK

For the current work we have used two most popular scripts, viz., *Bangla* and *Devanagri*. Handwritten documents of both these scripts are found to contain many

English words (written in Roman script). Therefore we have designed two separate systems, one for script separation between *Devanagri* and English words, and the other between *Bangla* and English words. For this purpose, we have collected text document written in *Bangla* and *Devanagri* scripts mixed with English words. The document pages are binarized using a simple adaptive threshold-based technique. Then we have applied our previously published Neighboring Component Analysis technique [1] to extract the text lines from the document pages and subsequently localize words from each line. After extracting the words, we have estimated 8 different word-level holistic features from these words. Two different MLP based classifiers are trained with the designed feature set to distinguish between *Devanagri*/Roman and *Bangla*/Roman scripts words. The key modules of the present work are shown in Fig. 2.

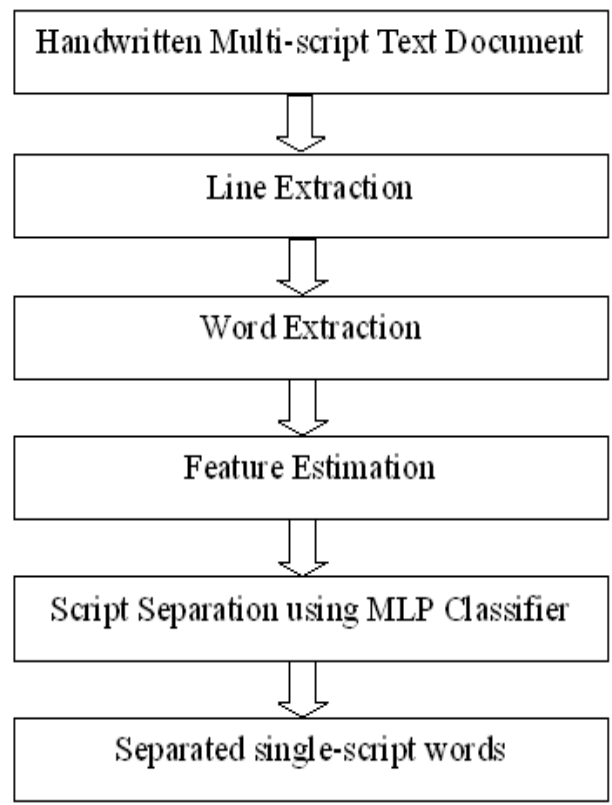

Fig 2: Key modules of the developed methodology

## 3 DATA ACQUISITION

The document pages for the database have been collected from three different types of sources, viz., class-notes of students, handwritten manuscripts of articles and from document pages written by different persons on request for the current project under our supervision. The writers (for the third category) were asked to use a black or a blue ink pen and write in *Bangla*/*Devanagri* or Roman scripts inside A-4 size pages. There were not any other restrictions regarding the content of the text.

The document pages were scanned at 300 dpi resolutions and were binarized through simple adaptive thre-sholding, where the threshold was chosen as the mean of the maximum and minimum gray level values in each document image. All the binarized images have been archived in a standard DAT format (where the foreground and background pixels are represented as '0' and '1' respectively). Then the documents were preprocessed in order to remove all the remaining salt and pepper noises like long lines in the border zone(s). To remove discontinuity in the pixel-level, we have used erosion and dilation [3], two popularly used morphological operator in image processing. Digitized document images are then categorized into two datasets. The Dataset#1 contains document image with both *Bangla* and Roman scripts and Dataset#2 contains document images with *Devanagri* and Roman scripts. Sample images from both these datasets are available in www.cmaterju.org.

## 4  LINE AND WORD EXTRACTION

In one of our earlier works [1] a script independent text line identification technique was reported. The work used a novel approach of height-specific dimensional comparison of neighborhood components to classify them into text lines. A post-processing step, based on Euclidean distance metric, efficiently classifies small components into text lines to which they most suitably belong. Large components suspected to be parts of two or more touching lines are also allocated to text lines efficiently. The method works well for unconstrained handwritten documents in which text lines overlap each other's bounding boxes to a great extent. With minimal data loss, the technique enhances the functionality of subsequent stages of an OCR system with multi-script handwritten texts. The word segmentation technique was based on projection profiles and used vertical pixel density histograms to effectively classify connected components into words. The word segmentation algorithm works well for both cursive and non-cursive handwriting.

## 5  DESIGN OF FEATURE SET

For any successful pattern classification system, it is very challenging and essential to select or estimate the features which are strong enough to categorize the input patterns to the proper output class to which they belong to. In our present work of script identification, we have designed 8 different features for the separation between *Devanagri*/Roman and *Bangla*/Roman scripts using MLP classifier. We have used these feature values with suitable normalization. The features are described as follows:

**Horizontalness feature:** In case of *Bangla* or *Devanagri* scripts, there is a prominent feature called Matra or headline which appears at the top of most characters. Generally, this Matra connects characters in a word. If horizontal pixel density is taken, it may be found that a steeper peak is found near the middle of the upper half of the word image. But in case of English word, there is no Matra like feature which distinguishes it from *Bangla* or *Devanagri* words. This horizontalness property of the Matra is extracted from the row wise sum of continuous run of black pixels, as discussed in our earlier work [2]. This is also illustrated in Fig. 3.

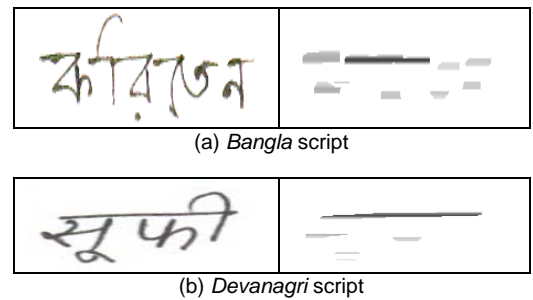

(a) *Bangla* script

(b) *Devanagri* script

Fig. 3 (a-b): An Illustration of Horizontalness feature

**Segmentation-based feature:** Here we have considered two features, viz., number of Matra pixels and number of segmentation-point pixels. We assumed that the words were segmented into constituent characters on the basis of Matra-based segmentation technique [9], generating Matra pixels and segmentation-point pixels. These pixels counts are used as two different feature values in the current work. Since English words doesn't have Matra region, these pixels counts are significuntllay lower in case of Roman script in caomparison to the words written in *Bangla* or *Devanagri* scripts. Fig. 4 shows a sample word image written in a Matra based (*Bangla*) script. Potential Matra pixels and segmentation-point pixels, obtained by applying the segmentation algorithm developed in [9], are also highlighted in the said figure.

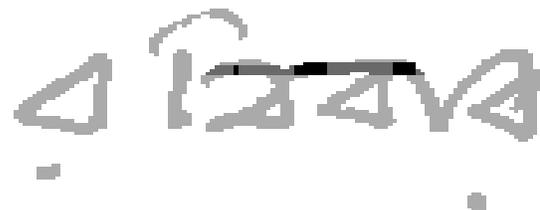

Fig. 4: Illustration of Matra pixels (medium grey) and segmentation-point pixels (dark grey) are shown on a sample *Bangla* word image in different grey shades.

**Foregroung-background Transition feature:** Analyzing the *Bangla*/*Devanagri* and English words it is often found that the horizontal pixel density of these words show different characteristics in different region of the word image. Considering this, we have taken the changeover of foreground and background pixels (transition point count) as feature value along 5 row positions where *Bangla*, *Devanagri* or English words appear to be differ most. For this purpose, we have made some assumptions like, the top row and bottom row of the word image are selected as R1 and R5 respectively. The row with maximum horizontalness is selected as R2. Estimation of R1, R2 and R5 are described in our earlier work on character segmentation [2]. Then we have specified 5 different rows such as, R2, R4=(R2+R5)/4, R3=(R2+R4)/3, R12=(R1+R2)/2 and R13=(R12+R2)/2. Finnaly, we have calculated the foreground-background feature values on these specific row positions. These rows are illustrated in Fig 5.

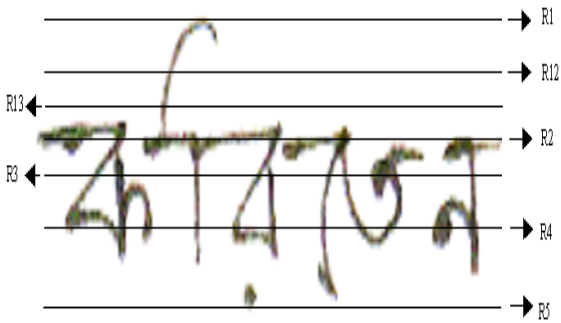

Fig. 5: Selection of specific rows for estimation of foreground-background feature values.

# 6  DESIGN OF PATTERN CLASSIFIER

In the present work, MLP classifiers are used for script classification of words generated from the document page. One classifier separates between *Devanagri*/Roman words where the other classifier distinguishes between *Bangla*/Roman scripts words.

The MLP classifier designed for this work is trained with the Back Propagation (BP) algorithm. It minimizes the sum of the squared errors for the training samples by conducting a gradient descent search in the weight space. The number of neurons in a hidden layer in the same is also adjusted during its training.

# 7  EXPERIMENTAL RESULTS

For developing a training set and a test set for each of the MLP based classifiers, employed for this work, the relevant database is divided in a ratio of 9:7. BP learning algorithm with learning rate $(\eta)$ = 0.8 and momentum term $(\alpha)$ = 0.8 is used here for training of these MLP based classifiers.

For classification of the words into 2 classes, the 8-element feature set, as discussed earlier, is used for both the datasets. The single layer MLP based classifier is designed for dataset#1 which consists of 8 neurons in input layer, 12 neurons in its hidden layer, and 2 neurons in its output layer i.e. the configuration is 8-12-2 and for dataset#2 the configuration is 8-16-2 i.e., 8 neurons are used in input layer, 16 neurons are used in hidden layer and 2 neurons are used in output layer. It has been observed empirically that, the best recognition rate on test set of dataset#1 and dataset#2 are achieved for 12 neurons and 16 neurons in the hidden layer respectively. For both datasets 450 and 350 sample patterns are considered per class for training and test purposes respectively. Being trained for 2000 iterations, it shows recognition performances of 99.29% and 98.43% on the test sets with *Bangla*-English words and *Devanagri*-English words respectively. The performance of the developed technique is also described in Table 1.

TABLE 1
PERFORMANCE DESCRIPTION OF THE DEVELOPED TECHNIQUE

|  | Number of words Trained | Number of words Tested | Successful script Separation |
| --- | --- | --- | --- |
| Dataset #1 | 900 | 700 | 99.29% |
| Dataset #2 | 900 | 700 | 98.43% |

Fig 6(a-c) shows sample word images from the both datasets which are separated successfully. Fig 7(a-c) illustrates sample images where the current technique fails to identify the script of the words correctly. As, in our methodology some of the features are Matra-based, therefore if any word is not having any Matra or there is discontinuity in the Matra then it may be classified incorrectly as shown in Fig 7(a). Also for the same reason if English word is having any Matra like component it may be misclassified as shown in Fig 7(b).

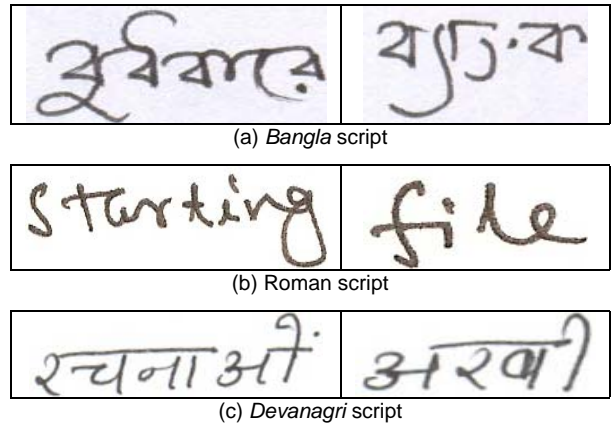

(a) *Bangla* script

(b) Roman script

(c) *Devanagri* script

Fig. 6 (a-c): Sample images of successful separation of different scripts

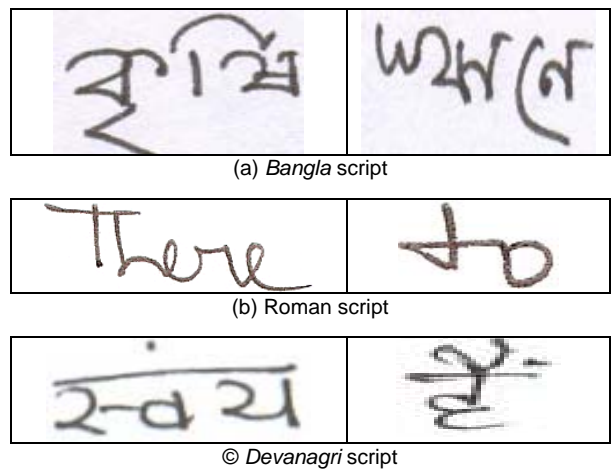

(a) *Bangla* script

(b) Roman script

© *Devanagri* script

Fig. 7 (a-c): Sample images of unsuccessful separation of different scripts

# 8  CONCLUSION

In the current work, we have developed a simple yet effective script separation methodology for *Bangla* and *Devanagri* handwritten words mixed with Roman scripts.

Not many prior works are available for word-level script separation from unconstrained handwritten Indic script documents. Because of long colonial past, presence of Roman script is common in Indic script documents. *Devanagri* and *Bangla*, two most popularly used scripts in Indian subcontinent, are no exception to that. The designed of the feature set is a novelty in the current work. Performances of the feature descriptor, on training with MLP based classifiers, on two different datasets are found to be comparable with the available state-of-the-art. This work may further be extended for word-level script separation involving different other popular Indic scripts. More data samples are to be collected for detailed evaluation of the developed methodology. Finally the designed technique is to be integrated with the proposed system for developing effective multi-lingual character recognition software.

**Acknowledgement.** Authors are thankful to the Center for Microprocessor Application for Training Education and Research (CMATER) and Project on Storage Retrieval and Understanding of Video for Multimedia (SRUVM) of Computer Science & Engineering Department, Jadavpur University, for providing infrastructure facilities during progress of the work. Also a lot of people helped us to make the database completion successfully. Authors are grateful to everyone who contributed with data to make this project successful.

**Ram Sarkar** received his B.Tech degree in Computer Science and Engineering from University of Calcutta, in 2003. He received his M.C.S.E degree from Jadavpur University (J.U.), in 2005. He joined J.U. as a lecturer in 2008. His areas of current research interest are document image processing, line extraction and segmentation of handwritten text images.

**Nibaran Das** received his B.Tech degree in Computer Science and Technology from Kalyani Govt. Engineering College under Kalyani University, in 2003. He received his M.C.S.E degree from Jadavpur University, in 2005. He joined J.U. as a lecturer in 2006. His areas of current research interest are OCR of handwritten text, Bengali fonts, biometrics and image processing. He has been an editor of Bengali monthly magazine "Computer Jagat" since 2005.

**Subhadip Basu** received his B.E. degree in Computer Science and Engineering from Kuvempu University, Karnataka, India, in 1999. He received his Ph.D. (Engg.) degree thereafter from Jadavpur University in 2006. He joined J.U. as a senior lecturer in 2006. His areas of current research interest are OCR of handwritten text, gesture recognition, real-time image processing.

**Mita Nasipuri** received her B.E.Tel.E., M.E.Tel.E., and Ph.D. (Engg.) degrees from Jadavpur University, in 1979, 1981 and 1990, respectively. Prof. Nasipuri has been a faculty member of J.U since 1987. Her current research interest includes image processing, pattern recognition, and multimedia systems. She is a senior member of the IEEE, U.S.A., Fellow of I.E (India) and W.B.A.S.T, Kolkata, India.

**Dipak Kumar Basu** received his B.E.Tel.E., M.E.Tel., and Ph.D. (Engg.) degrees from Jadavpur University, in 1964, 1966 and 1969

respectively. Prof. Basu has been a faculty member of J.U from 1968 to January 2008. He is presently an A.I.C.T.E. Emiretus Fellow at the CSE Department of J.U. His current fields of research interest include pattern recognition, image processing, and multimedia systems. He is a senior member of the IEEE, U.S.A., Fellow of I.E. (India) and W.B.A.S.T., Kolkata, India and a former Fellow, Alexander von Humboldt Foundation, Germany.